\documentclass[runningheads]{llncs}
\usepackage{graphicx}

\usepackage{tikz}
\usepackage{comment}
\usepackage{amsmath,amssymb}
\usepackage{color}
\usepackage{xcolor}
\usepackage{multirow}
\usepackage{tabularx}
\usepackage{pifont}
\usepackage[colorlinks]{hyperref}
\usepackage[accsupp]{axessibility}

\newcommand{\cmark}{\ding{51}}%
\newcommand{\xmark}{\ding{55}}%

\newcolumntype{L}[1]{>{\raggedright\let\newline\\\arraybackslash\hspace{0pt}}m{#1}}
\newcolumntype{C}[1]{>{\centering\let\newline\\\arraybackslash\hspace{0pt}}m{#1}}
\newcolumntype{R}[1]{>{\raggedleft\let\newline\\\arraybackslash\hspace{0pt}}m{#1}}

\makeatletter
\newcommand*{\rom}[1]{\expandafter\romannumeral #1}
\def\blfootnote{\xdef\@thefnmark{}\@footnotetext}
\makeatother

\begin{document}
\pagestyle{headings}
\mainmatter
\def\ECCVSubNumber{5036}  
    
\title{3D Clothed Human Reconstruction in the Wild} 

\titlerunning{3D Clothed Human Reconstruction in the Wild}

\author{
Gyeongsik Moon\inst{1}$^{*}$ \and
Hyeongjin Nam\inst{2}$^{*}$ \and
Takaaki Shiratori\inst{1} \and \\
Kyoung Mu Lee\inst{2,3}
}

\authorrunning{G. Moon et al.}

\institute{
Meta Reality Labs Research \and
Dept. of ECE \& ASRI, Seoul National University, Korea \and
IPAI, Seoul National University, Korea \\
\email{\{mks0601,tshiratori\}@fb.com}, \email{\{namhjsnu28,kyoungmu\}@snu.ac.kr}
}
\maketitle

\begin{abstract}
Although much progress has been made in 3D clothed human reconstruction, most of the existing methods fail to produce robust results from in-the-wild images, which contain diverse human poses and appearances.
This is mainly due to the large domain gap between training datasets and in-the-wild datasets.
The training datasets are usually synthetic ones, which contain rendered images from GT 3D scans.
However, such datasets contain simple human poses and less natural image appearances compared to those of real in-the-wild datasets, which makes generalization of it to in-the-wild images extremely challenging.
To resolve this issue, in this work, we propose ClothWild, a 3D clothed human reconstruction framework that firstly addresses the robustness on in-the-wild images.
First, for the robustness to the domain gap, we propose a weakly supervised pipeline that is trainable with 2D supervision targets of in-the-wild datasets.
Second, we design a DensePose-based loss function to reduce ambiguities of the weak supervision. 
Extensive empirical tests on several public in-the-wild datasets demonstrate that our proposed ClothWild produces much more accurate and robust results than the state-of-the-art methods.
The codes are available in \href{https://github.com/hygenie1228/ClothWild_RELEASE}{here}.
\blfootnote{* equal contribution}
\blfootnote{This work was primarily done while Gyeongsik Moon was in SNU.}

\end{abstract}

\section{Introduction}
\label{sec:introduction}
3D clothed human reconstruction aims to reconstruct humans with various poses, body shapes, and clothes in 3D space from a single image.
It is an essential task for various applications, such as 3D avatars and virtual try-on.
Most of the recent 3D clothed human reconstruction methods~\cite{saito2019pifu,zheng2019deephuman,alldieck2019tex2shape,saito2020pifuhd,huang2020arch,he2020geo,he2021arch++,bhatnagar2019multi,jiang2020bcnet,corona2021smplicit,alldieck2022photorealistic} require 3D scans for the training; hence, they are trained on synthetic datasets, which consist of 3D scans~\cite{renderpeople,axyz} and rendered images from the scans.
Although significant progress has been made by utilizing such synthetic datasets, all of them fail to produce robust results on in-the-wild images.

\begin{figure}[t]
\centering
\includegraphics[width=1.0\linewidth]{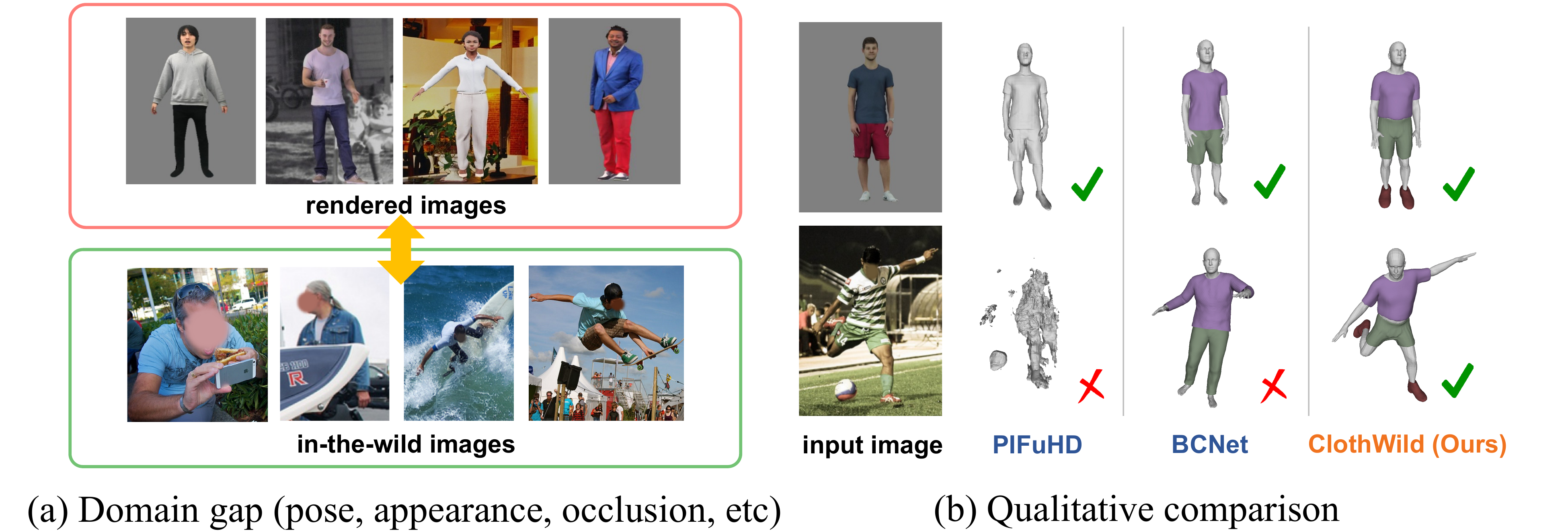}
   \caption{
   (a) Domain gap between rendered images and in-the-wild images.
   (b) Due to the large domain gap, existing methods (\textit{e.g.} PIFuHD~\cite{saito2020pifuhd} and BCNet~\cite{jiang2020bcnet}) fail on in-the-wild images, while our ClothWild successfully reconstructs. 
   Colors of reconstructed 3D clothes are manually assigned to represent cloth types.
   }   
\label{fig:introduction}
\end{figure}

The in-the-wild images are taken in our daily environments, such as cluttered offices and concert halls, and have diverse human poses, appearances, and severe human occlusions.
On the other hand, the 3D scans are captured from restricted environments, such as motion capture studios.
To ease 3D scan capturing, subjects are often instructed to take simple poses.
Therefore, rendered images from the 3D scans have artificial appearances and simple human poses compared to those of in-the-wild images, as shown in Fig.~\ref{fig:introduction}(a).
This large domain gap between in-the-wild and rendered images makes the methods trained on synthetic datasets fail to generalize to in-the-wild images, as shown in Fig.~\ref{fig:introduction}(b).

For the robustness to the domain gap, we present ClothWild, a 3D clothed human reconstruction framework that leverages a weak-supervision strategy to learn 3D clothed humans from in-the-wild datasets.
Here, the weak supervision means that the supervision target is not full 3D data, but 2D data defined only in the single-camera viewpoint (\textit{i.e.}, 2D cloth segmentations~\cite{gong2017look}).
For the weak-supervision strategy, our reconstruction pipeline is divided into two networks: a cloth generative model and a regressor, ClothNet.
The cloth generative model generates 3D cloth geometry around the T-posed 3D body of SMPL~\cite{loper2015smpl} from the cloth latent codes, where the 3D cloth geometry is animatable using a skinning function of SMPL.
It is fully supervised on the synthetic datasets beforehand, and we freeze it during the weakly supervised training.
Although the cloth generative model covers various cloth types and styles, it does not cover complicated human poses and diverse image appearances of in-the-wild datasets.
For the robustness to such in-the-wild datasets, we design a regressor, called ClothNet, and weakly supervise it on in-the-wild datasets to predict latent codes of the cloth generative model.
The final output can be obtained by passing the output of the ClothNet to the cloth generative model.
As the cloth generative model is fixed during the weakly supervised training, we can protect the generative model from being hurt by imperfect 2D supervision targets of in-the-wild datasets, while making our ClothNet robust to diverse human poses and image appearances of in-the-wild images.
Our weak-supervision strategy can recover full 3D geometry only from the evidence of the single-camera viewpoint by regularizing the predicted cloth latent codes to be in the cloth latent space.
Note that such regularization is not possible for previous approaches~\cite{saito2019pifu,saito2020pifuhd,xiu2022icon,alldieck2022photorealistic} as they do not model cloth latent space.

For the weak supervision, we can enforce the projection of 3D reconstruction results to be close to 2D supervision targets (\textit{i.e.}, cloth segmentations~\cite{gong2017look}) of in-the-wild datasets.
However, naively adopting such a strategy has three difficulties in learning 3D clothed humans.
First, the 2D supervision targets provide information for only a single camera viewpoint.
Therefore, there is no guarantee that the weak supervision can recover 3D geometry of other viewpoints.
Second, severe depth ambiguity of the cloth segmentations hampers learning accurate 3D clothes.
Since each pixel in cloth segmentations corresponds to innumerable points in 3D space, it is difficult to specify 3D points corresponding to clothes.
Third, pixel-level misalignment can occur between projected 3D reconstruction results and 2D cloth segmentation.
As camera parameters are not available for in-the-wild images, we should additionally predict the camera parameters to project the 3D clothed humans.
This is a highly ill-posed problem as various combinations of camera parameters and 3D clothed humans correspond to the same 2D cloth segmentation.
Due to the ill-posedness, camera prediction can often fail, which results in the wrong projection to the 2D image space.

In this regard, we design a DensePose-based loss function to resolve the issues which can occur when weakly supervising 3D clothed human reconstructions based on their projections.
DensePose~\cite{guler2018densepose} informs where each human pixel corresponds to the 3D human body surface of SMPL.
Using DensePose, our DensePose-based loss function obtains 3D points that correspond to the cloth segmentations around the SMPL surface.
Then, it enforces such 3D points to be a part of the 3D cloth surface.
As the DensePose effectively limits possible positions of 3D points around the SMPL surface, it significantly reduces the depth ambiguity.
In addition, DensePose is already aligned with cloth segmentations in the image space; therefore, we do not have to predict camera parameters and project 3D reconstructions to the 2D image space.
Hence, our DensePose-based loss function does not suffer from pixel-level misalignment.

We show that the proposed ClothWild produces far more robust and better results than the previous 3D clothed human reconstruction methods from in-the-wild images.
As robustness on in-the-wild images has not been extensively studied in the 3D human reconstruction community, we believe ours can give useful insights to the following research.

Our contributions can be summarized as follows.
\begin{itemize}
\item We present ClothWild, which reconstructs robust 3D clothed humans from a single in-the-wild image. 
To the best of our knowledge, it is the first to explicitly address robustness on in-the-wild images.
\item For the robustness to the domain gap between synthesized and in-the-wild datasets, we propose a weakly supervised pipeline that is trainable with 2D supervision targets of in-the-wild datasets.
\item Our DensePose-based loss function resolves the ambiguities of weak supervision by effectively limiting possible positions of 3D points using DensePose.
\item ClothWild largely outperforms previous methods on in-the-wild images.
\end{itemize}
\section{Related works}
\noindent\textbf{3D clothed human reconstruction.} 
Varol~et al.~\cite{varol2018bodynet} and Jackson~et al.~\cite{jackson20183d} proposed volumetric regression networks that directly predict a voxel representation of a 3D clothed human from a single image.
Saito~et al.~\cite{saito2019pifu,saito2020pifuhd} and He~et al.~\cite{he2020geo} presented a pixel-aligned implicit function that predicts 3D occupancy fields of clothed humans.
Alldieck~et al.~\cite{alldieck2019learning} utilized displacement vectors between a human body surface and clothes for reconstruction.
Alldieck~et al.~\cite{alldieck2019tex2shape} presented a method estimating normal and displacement maps on top of a human body model, such as SMPL~\cite{loper2015smpl}.
Bhatnagar~et al.~\cite{bhatnagar2019multi} and Jiang~et al.~\cite{jiang2020bcnet} presented 3D clothed human reconstruction systems that predict PCA coefficients of cloth generative model space.
Huang~et al.~\cite{huang2020arch} and He~et al.~\cite{he2021arch++} handle implicit function representation of 3D clothed humans in arbitrary poses and produce animatable reconstruction results.
Xiu~et al.~\cite{xiu2022icon} presented ICON, which utilizes local features for robust 3D clothed human reconstruction.
Alldieck~et al.~\cite{alldieck2022photorealistic} proposed PHORHUM to photorealistically reconstruct the 3D geometry and appearance of a dressed person.
As all of the above methods require 3D scans for training, they are trained on synthetic datasets, which consist of 3D scans and the rendered images from the scans.
As the synthetic datasets mainly contain simple human poses with artificial appearances, the methods trained on such datasets fail to generalize to in-the-wild datasets.
In contrast, our ClothWild is the first work that can be weakly supervised with 2D supervision targets of in-the-wild datasets, which results in robust outputs on in-the-wild images.

Recently, Corona~et al.~\cite{corona2021smplicit} presented a fitting framework that fits their cloth generative model to 2D cloth segmentations.
Although their fitting framework handles reconstruction on in-the-wild images, there are two major differences from ours.
First, their fitting framework is not a learning-based system; hence, it cannot utilize image features.
As it only relies on 2D cloth segmentations for fitting, it produces wrong results when cloth segmentations are not available due to truncations or occlusions.
On the other hand, ClothWild is a learning-based system that utilizes image features; hence, it is much more robust to occlusions and truncations by considering contextual information of image features, as shown in Fig.~\ref{fig:comparison_sota}.
Second, their fitting framework suffers from the depth ambiguity and pixel-misalignment as described in Section~\ref{sec:introduction}, since it projects 3D reconstruction results to the 2D image space and compares the projected one with 2D cloth segmentations.
On the other hand, ClothWild leverages our newly proposed DensePose-based loss function.
Hence, ClothWild suffers much less from the depth ambiguity and is free from the pixel-misalignment issue.
The detailed description of it is provided in Section~\ref{sec:ablation_densepose_loss}.

\noindent\textbf{3D cloth generative model.}
3D cloth generative models parameterize 3D clothes by embedding acquired 3D cloth scans in a latent space.
Bhatnagar~et al.~\cite{bhatnagar2019multi} and Jiang~et al.~\cite{jiang2020bcnet} used the PCA algorithm to represent clothes in a latent space.
Ma~et al.~\cite{ma2020learning} used a graph convolutional neural network-based generative model to embed the clothes in a latent space.
Bertiche~et al.~\cite{bertiche2020cloth3d} embedded displacements from the human body surface to registered cloth meshes and cloth types as latent codes.
Patel~et al.~\cite{patel2020tailornet} decomposed 3D clothes into low- and high-frequency components and represented them as a function of human poses, shapes and cloth styles.
Corona~et al.~\cite{corona2021smplicit} presented a cloth generative model, SMPLicit, which embeds 3D clothes as latent codes that represent cloth styles and cloth cuts.
SMPLicit covers a wide variety of clothes, which differ in their geometric properties, such as sleeve length and looseness.
In our framework, we employ SMPLicit as a cloth generative model.

\noindent\textbf{3D human body reconstruction.} 
3D human body reconstruction methods predict parameters of the SMPL~\cite{loper2015smpl} body model from a single image.
They perform well on in-the-wild images via weakly supervision with 2D GTs of in-the-wild images.
Kanazawa~et al.~\cite{kanazawa2018end} proposed an end-to-end framework that utilizes adversarial loss to reconstruct plausible human bodies.
Kolotouros~et al.~\cite{kolotouros2019learning} combined a regressor and iterative fitting framework.
Moon~et al.~\cite{Moon_2022_CVPRW_Hand4Whole} presented Pose2Pose utilizing local and global image features.
We use Pose2Pose as an off-the-shelf 3D human body reconstruction method of our ClothWild due to its superior accuracy compared to other works.

\section{ClothWild}
\begin{figure} [t]
\centering
\includegraphics[width=1.0\linewidth]{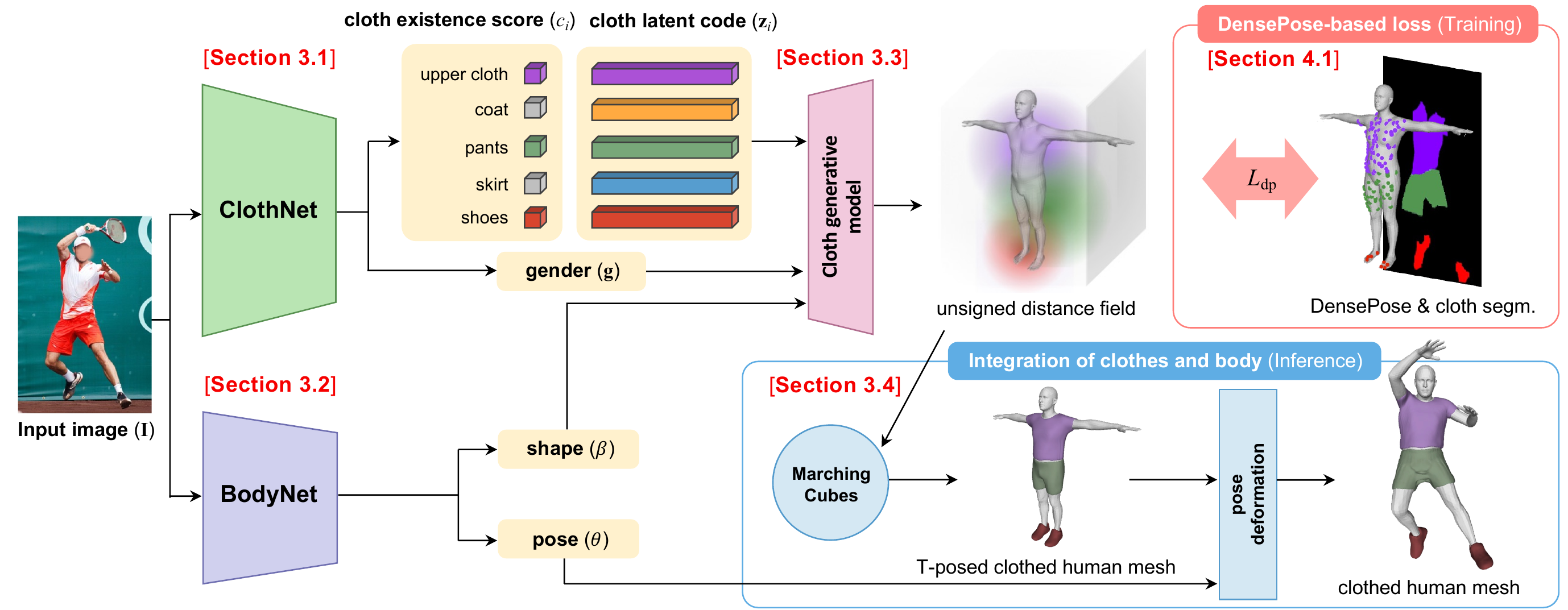}
   \caption{
   The overall pipeline of ClothWild.
   ClothNet predicts cloth latent codes of the cloth generative model, SMPLicit, and produces an unsigned distance field by passing the codes to the SMPLicit. 
   BodyNet predicts SMPL pose and shape parameters.
   At the training stage, the unsigned distance field is supervised with a combination of DensePose and cloth segmentations with our DensePose-based loss function.
   At the inference stage, the final 3D clothed human reconstruction is obtained through Marching Cubes and pose deformation steps.
   }
\label{fig:overall_pipeline}
\end{figure}

Fig.~\ref{fig:overall_pipeline} shows the overall pipeline of our ClothWild, which consists of ClothNet and BodyNet.
We provide a detailed description of each module below.

\subsection{ClothNet}
\label{sec:clothnet}
Given an input image $\mathbf{I}$, ClothNet predicts cloth existence scores $\mathbf{c}=(c_1, \ldots, c_{N_c})$, a set of cloth latent codes $\{\mathbf{z}_i\}_{i=1}^{N_c}$, and a gender $\mathbf{g}$.
$N_c = 5$ denotes the number of clothes we consider, which include upper cloth, coat, pants, skirt, and shoes.
The $i$th cloth existence score $c_i$ represents the probability that a human is wearing $i$th cloth.
The $i$th cloth latent code $\mathbf{z}_i \in \mathbb{R}^{d_i}$ represents a low-dimensional code of $i$th 3D cloth, embedded in latent space of the 3D cloth model, SMPLicit~\cite{corona2021smplicit}.
We set $d_i=4$ for shoes and $d_i=18$ for other clothes following SMPLicit~\cite{corona2021smplicit}.
The gender $\mathbf{g} \in \mathbb{R}^{2}$ is a one-hot encoded vector that represents a label for the appropriate human body model (\textit{i.e.}, male and female).
We use ResNet-50~\cite{he2016deep} to extract an image feature vector $\mathbf{f} \in \mathbb{R}^{2048}$ from the input image after removing the fully-connected layer of the last part of the original ResNet.
The image feature vector $\mathbf{f}$ is passed into a fully-connected layer, followed by a sigmoid activation function, to predict the cloth existence scores $\mathbf{c}$.
In addition, we use $N_c$ fully-connected layers to predict a set of cloth latent codes $\{\mathbf{z}_i\}_{i=1}^{N_c}$ from $\mathbf{f}$, where $i$th fully-connected layer predicts the latent codes of $i$th cloth, $\mathbf{z}_i$.
Finally, another fully-connected layer, followed by a softmax activation function, predicts the gender from $\mathbf{f}$.
The predicted set of cloth latent codes $\{\mathbf{z}_i\}_{i=1}^{N_c}$ and gender $\mathbf{g}$ are passed to SMPLicit.

\subsection{BodyNet}
\label{sec:bodynet}
The BodyNet takes the input image $\mathbf{I}$ and predicts a shape parameter $\beta \in \mathbb{R}^{10}$ and a pose parameter $\theta \in \mathbb{R}^{72}$ of the SMPL human body model~\cite{loper2015smpl}.
The shape parameter $\beta$ represents PCA coefficients of T-posed human body shape space, and the pose parameter $\theta$ represents 3D rotations of human body joints.
The shape parameter $\beta$ is forwarded to SMPLicit that is described in Section~\ref{sec:smplicit}.
The pose parameter $\theta$ is used in the inference stage, of which detailed descriptions are in Section~\ref{sec:integration}.
We use Pose2Pose~\cite{Moon_2022_CVPRW_Hand4Whole} as the BodyNet, which achieves state-of-the-art performance on in-the-wild benchmarks.

\subsection{Cloth generative model}
\label{sec:smplicit}
The cloth generative model embeds 3D clothes in the latent space.
We use SMPLicit~\cite{corona2021smplicit} as our cloth generative model, which produces a continuous unsigned distance field of a 3D cloth from a cloth latent code, gender, and human shape.
We use the pre-trained SMPLicit and fix it while training ClothWild.
Given the cloth latent codes $\{\mathbf{z}_i\}_{i=1}^{N_c}$, gender $\mathbf{g}$, and human shape $\beta$, SMPLicit outputs the unsigned distance field for the $i$th cloth, as follows:
\begin{equation}
\label{eq1}
C(\mathbf{x}, \mathbf{z}_{i}, \mathbf{g}, \beta) \longrightarrow  \mathbb{R}^+ , \quad  i = 1, 2, \ldots, N_c, 
\end{equation}
where $\mathbf{x} \in \mathbb{R}^{3}$ is a 3D query point in a canonical 3D space where the human is in a T-pose.
In the canonical 3D space, a T-posed naked human body mesh is founded by deriving from the gender $\mathbf{g}$ and human shape $\beta$.
Around the T-posed naked human body, a set of 3D query points indicate the closest distance to the 3D cloth surface as the unsigned distance field.
The above step only proceeds for the $i$th cloth when its cloth existence score $c_i$ is larger than a threshold, which we empirically set to 0.25.

\subsection{Integration of clothes and body}
\label{sec:integration}
\noindent\textbf{Marching Cubes.}
In the inference stage, we obtain cloth meshes using Marching Cubes~\cite{lorensen1987marching}, which extracts a mesh of an isosurface from a 3D discrete scalar field, such as the unsigned distance field.
We calculate the unsigned distance field by densely sampling 3D query points in the 3D space and forwarding them to the SMPLicit model.
Then, we extract cloth meshes on the T-pose from the unsigned distance field by Marching Cubes.
At the end of this step, we obtain a T-posed clothed human mesh that comprises a T-posed naked human body mesh and cloth meshes.

\noindent\textbf{Pose deformation.}
Finally, we apply the pose deformation to the naked human body mesh and cloth meshes of the T-posed clothed human mesh, respectively, following Corona~et al.~\cite{corona2021smplicit}.
To deform the naked human body, we use the skinning deformation of SMPL with the pose parameter $\theta$ predicted by the BodyNet.
To deform cloth meshes, we allocate each cloth vertex of the cloth meshes to its closest naked human body vertex and apply the same pose deformation of the human body vertex.
Such the SMPL-driven pose deformation has the strength that the reconstruction results can be deformed with an arbitrary pose instead of the predicted pose, which enables an animation.

\section{Learning from in-the-wild datasets}
\label{sec:learning}
In this section, we describe loss functions to train our framework on in-the-wild datasets without any 3D scan GTs.
ClothNet is the only trainable module in our framework, and all other modules, including BodyNet and SMPLicit, are fixed during the training.
The overall loss function is
\begin{equation}
L_{\text{total}} = 
\lambda_{\text{dp}}L_{\text{dp}}+\lambda_{\text{reg}}L_{\text{reg}}+\lambda_{\text{exist}}L_{\text{exist}}+\lambda_{\text{gender}}L_{\text{gender}}, 
\end{equation}
where $\lambda_{\text{dp}}=1$, $\lambda_{\text{reg}}=0.1$, $\lambda_{\text{exist}}=0.01$, and $\lambda_{\text{gender}}=0.01$.

\subsection{DensePose-based loss function}
\label{sec:dp_loss}
The DensePose-based loss function $L_{\text{dp}}$, which uses a combination of cloth segmentations and DensePose, makes reconstructed clothes close to cloth segmentations.
The loss is computed in three steps: cloth-to-body mapping, query point selection, and loss calculation.

\begin{figure}[t]
\centering
\includegraphics[width=0.8\linewidth]{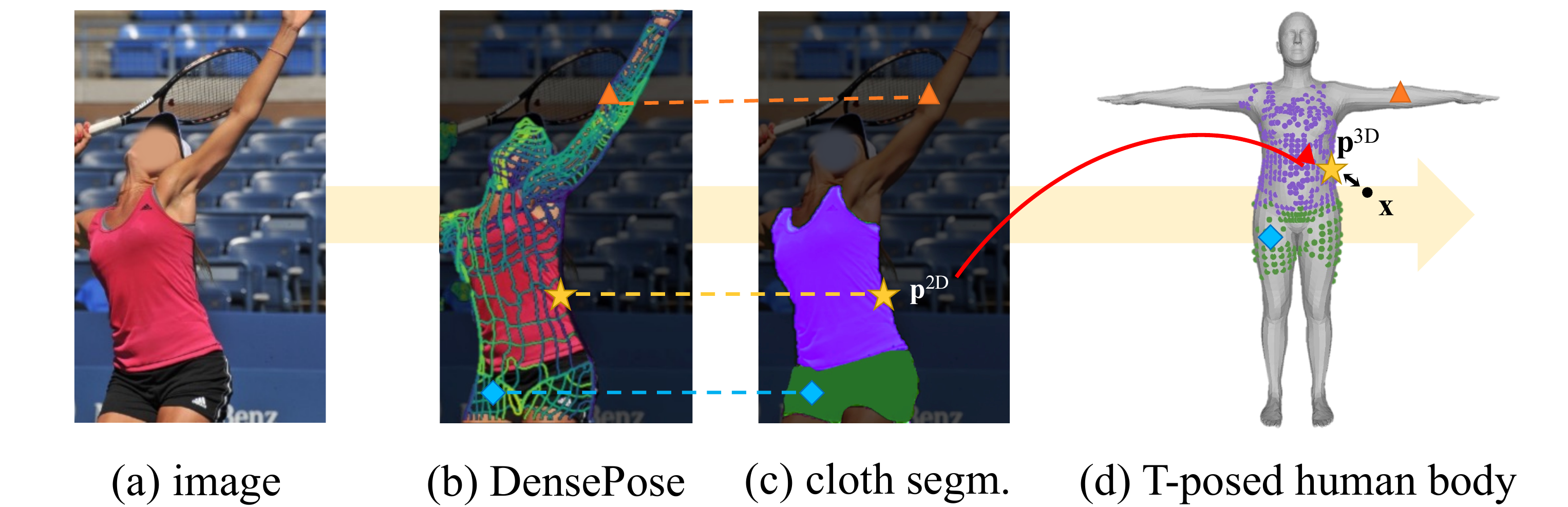}
   \caption{
   The procedure of the cloth-to-body mapping step.
   From (a) an in-the-wild image, we exploit 2D supervision targets: (b) DensePose and (c) cloth segmentations.
   (d) We sample a 2D cloth point $\mathbf{p}^{\text{2D}}$ of the cloth segmentations, and map the 2D cloth point $\mathbf{p}^{\text{2D}}$ to the 3D cloth point $\mathbf{p}^{\text{3D}}$ on the T-posed human body surface.
   We select the 3D query point $\mathbf{x}$ within constant distance from the 3D cloth points, and the DensePose-based loss function supervises the selected 3D query point.
   }
\label{fig:loss_description}
\end{figure}

\noindent\textbf{Cloth-to-body mapping.} 
In the first step, we place pixels of 2D cloth segmentations in 3D space.
Fig.~\ref{fig:loss_description} shows the procedure of the cloth-to-body mapping step.
In the area where DensePose is defined, we sample 2D cloth points $\{\mathbf{p}^{\text{2D}}_{k}\}_{k=1}^{N_{p}}$ on cloth segmentations, where $N_{p}=196$ is the number of the sampled 2D points.
We utilize DensePose that informs where each human pixel corresponds to the 3D human body surface of SMPL to map the 2D cloth points $\{\mathbf{p}^{\text{2D}}_{k}\}_{k=1}^{N_{p}}$ to 3D cloth points $\{\mathbf{p}^{\text{3D}}_{k}\}_{k=1}^{N_{p}}$ on the T-posed human body surface.
Each of the mapped 3D cloth points represents a cloth label (\textit{e.g.}, upper cloth, pants, or non-cloth) elicited from the cloth segmentations.

\noindent\textbf{Query point selection.}
In this step, we select 3D query points around the T-posed human body for loss calculation.
For each cloth type, we uniformly sample 3D query points at a resolution of $21\times21\times21$ from a 3D bounding box, where the sampling strategy is described in the supplementary material.
We select 3D query points $\{\mathbf{x}_{j}\}_{j=1}^{N_{q}}$ within a distance threshold $\tau$ from the 3D cloth points $\{\mathbf{p}^{\text{3D}}_{k}\}_{k=1}^{N_{p}}$ on the 3D human body surface, where $N_{q}$ is the number of the sampled 3D query points.
The distance threshold $\tau$ is set 10 cm for the coat and 3 cm for others.

\noindent\textbf{Loss calculation.}
To the formal description of the loss function, we define a cloth binary function $S_{i}(\cdot)$.
If the closest 3D cloth point from a 3D query point $\mathbf{x}_{j}$ belongs to the $i$th cloth, $S_{i}(\textbf{x}_{j})$ becomes 1 and 0 else.
With the cloth binary function $S_{i}(\cdot)$, the DensePose-based loss function follows:
\begin{equation}
\begin{split}
L_{\text{dp}} = 
{1 \over N_{c}N_{q}}\sum\limits_{i=1}^{N_{c}}\sum\limits_{j=1}^{N_{q}} (
S_{i}(\mathbf{x}_{j})|C(\mathbf{x}_{j}, \mathbf{z}_{i}, \mathbf{g}, \beta)| \\ 
+(1-S_{i}(\mathbf{x}_{j}))|C(\mathbf{x}_{j}, \mathbf{z}_{i}, \mathbf{g}, \beta)-d_{\text{max}}| ),
\end{split}
\end{equation}
where $d_{\text{max}}$ is a maximum cut-off distance of the unsigned distance fields, set to 0.01 for shoes and 0.1 for others.
The first loss term forces the 3D query point $\mathbf{x}$ to belong to $i$th cloth by making its unsigned distance close to zero when the 3D point matches cloth label $i$.
The second loss term forces the 3D query point $\mathbf{x}$ to not belong to $i$th cloth when the 3D point does not match cloth label $i$.
As a result, the DensePose-based loss function supervises the 3D query points established by the above steps without the depth ambiguity of the cloth segmentations.

\subsection{Other loss functions}
\label{other_loss}
\noindent\textbf{Regularization loss.} 
The regularization loss function $L_{\text{reg}}$ makes predicted latent codes close to the mean of the cloth latent space, which results in plausible 3D clothes.
The regularization loss function is defined as follows:
$L_{\text{reg}} = 
\sum_{i=1}^{N_c} \alpha_{i} \lVert \mathbf{z}_{i} \rVert_{2}$, where $\alpha_{i}$ is set 0.1 for shoes and 1.0 for others.
As the DensePose-based loss supervises 3D points that belong to partial areas of clothes, there is a possibility to learn implausible clothes (\textit{e.g.}, overly thick cloth or torn cloth) that only rely on the areas.
The regularization loss prevents it by constraining output clothes to be close to the mean.

\noindent\textbf{Cloth existence loss.}
For the cloth existence prediction, we calculate a binary cross-entropy as follows: 
$L_{\text{exist}} = 
-{1 \over N_{c}}\sum_{i=1}^{N_c} (c^*_i \text{log}c_i + (1-c^*_i)(1-\text{log}c_i))$, where the asterisk denotes the groundtruth.

\noindent\textbf{Gender classification loss.} 
For the gender classification, we calculate a cross-entropy as follows:
$L_{\text{gender}} = 
-(g^*_m \text{log}g_m + g^*_f \text{log}g_f)$
, where $g_m$ and $g_f=1-g_m$ is the probability of being male and female of the input human, respectively.
The asterisk denotes the groundtruth.
\section{Implementation details}
PyTorch~\cite{paszke2017automatic} is used for implementation. 
The backbone part is initialized with the publicly released ResNet50~\cite{he2016deep} pre-trained on ImageNet~\cite{russakovsky2015imagenet}. 
The weights are updated by Adam optimizer~\cite{kingma2014adam} with a mini-batch size of 8. 
The human body region is cropped using a GT box in both training and testing stages following previous works~\cite{saito2019pifu,saito2020pifuhd}.
The cropped image is resized to 256$\times$192.
Data augmentations, including scaling, rotation, random horizontal flip, and color jittering, are performed in training.
The initial learning rate is set to $10^{-4}$ and reduced by a factor of 10 after the 5th epoch.
We train the model for 8 epochs with an NVIDIA GTX 2080 Ti GPU. 
\section{Experiment}
\subsection{Datasets}
\noindent\textbf{MSCOCO.}
MSCOCO~\cite{lin2014microsoft} is a large-scale in-the-wild dataset, which provides in-the-wild images with diverse human poses and appearances.
We train ours on the training split and evaluate on the validation split.
We use outputs of the DensePose regression model~\cite{guler2018densepose} for the DensePose-based loss function as GT DensePose provides only sparse annotations.
For the cloth segmentations, we use LIP~\cite{gong2017look}, which contains cloth segmentation annotations of MSCOCO images.
We acquire gender annotations by running Homogenus~\cite{pavlakos2019expressive} on MSCOCO and use its predictions as supervision targets.

\noindent\textbf{DeepFashion2.}
DeepFashion2~\cite{ge2019deepfashion2} is a comprehensive dataset that focuses on capturing clothed humans with a wide variety of cloth styles, and we use it as the additional train set along with the MSCOCO.
We obtain DensePose and gender annotations by the same procedure as MSCOCO, described above.
For cloth segmentation annotations, we use SCHP~\cite{li2020self} to obtain cloth segmentations of the dataset.

\noindent\textbf{3DPW.}
3DPW~\cite{von2018recovering} is an in-the-wild dataset, and we use it only for the evaluation purpose after sampling every 25th frame of test set videos.
We use two items of 3DPW: T-posed clothed human meshes registered to each subject's 3D scan and SMPL pose parameters of humans in images.
As the registered 3D clothed human meshes have the same mesh topology as that of SMPL body mesh, we deform the registered ones with SMPL pose parameters following the same skinning algorithm of SMPL.
We use the posed registered meshes as GT 3D clothed humans of 3DPW.

\begin{figure}[t]
\centering
\includegraphics[width=\linewidth]{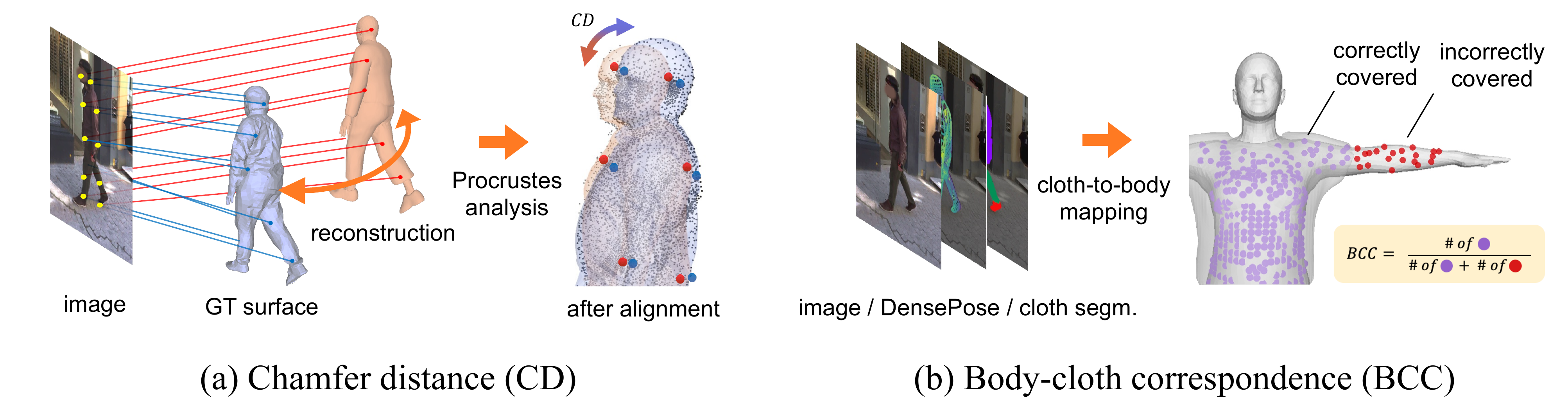}
   \caption{
    Descriptions of our evaluation metrics.
    (a) Chamfer distance (CD) is a 3D distance between reconstruction and GT surface.
    It is measured after aligning them based on each 2D projection.
    (b) BCC is a proportion of 3D points that have correctly matched cloth types.
    The purple and red points on the right body surface are 3D points that are mapped from cloth segmentations using DensePose.
    The purple points represent 3D points that are covered with a correct type of reconstructed 3D clothes, while the red points represent others.
    It is measured after normalizing 3D poses.
       }
\label{fig:metric_description}
\end{figure}

\subsection{Evaluation metrics}
Fig.~\ref{fig:metric_description} shows our two evaluation metrics.
Since there has been no work dealing with 3D clothed human reconstruction on the in-the-wild dataset, we propose two evaluation metrics in the following.

\begin{figure}[t]
\centering
\includegraphics[width=\linewidth]{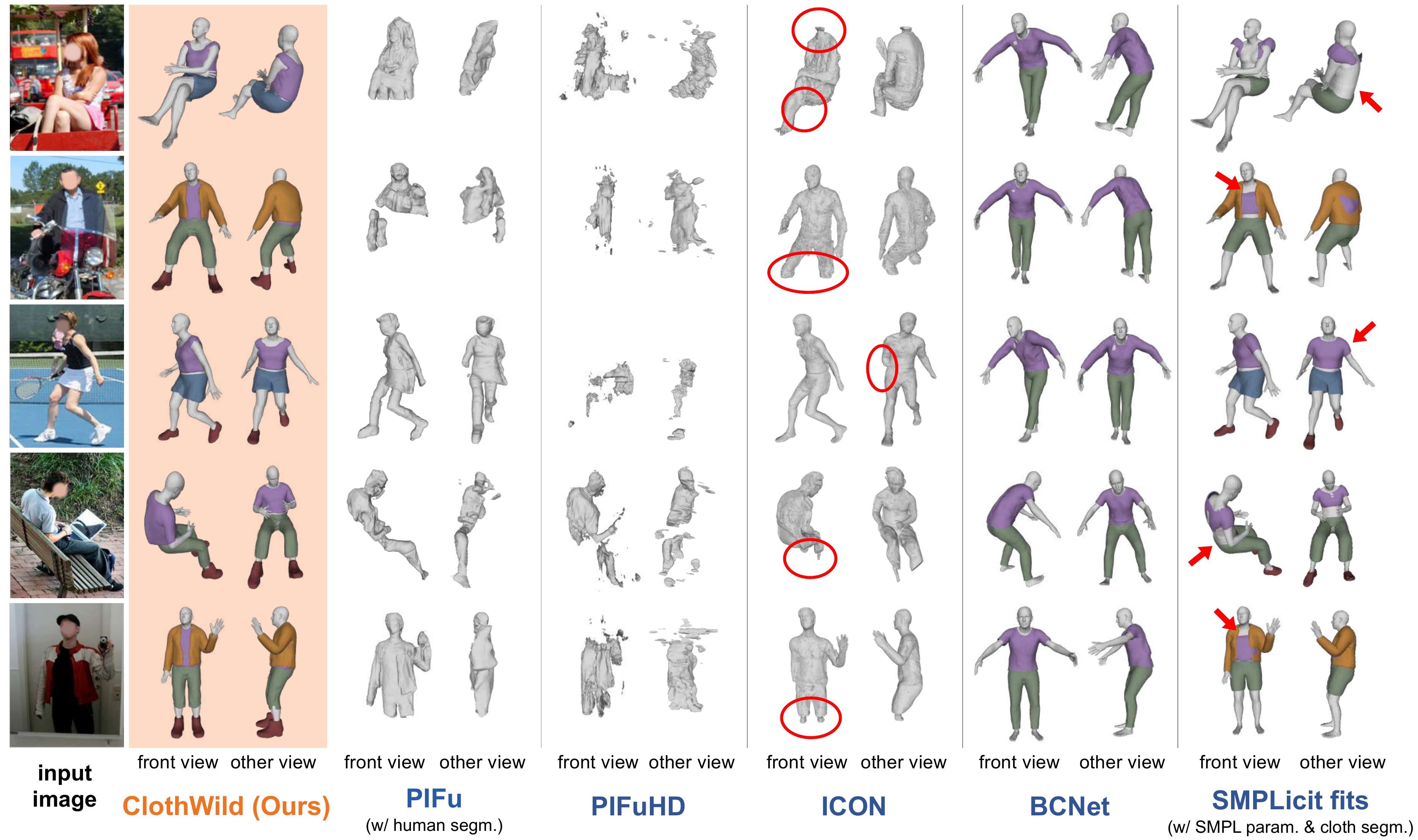}
\caption{
Qualitative result comparisons on MSCOCO validation set.
PIFu additionally uses human segmentation obtained from Mask R-CNN~\cite{he2017mask} for reconstruction.
SMPLicit fits use SMPL parameter and cloth segmentations obtained from our BodyNet and SCHP~\cite{li2020self}, respectively. 
Colors of reconstructed 3D clothes are manually assigned to represent cloth types.
}
\label{fig:comparison_sota}
\end{figure}

\noindent\textbf{Chamfer distance (CD).}
Chamfer Distance (CD) is a 3D distance between the 3D clothed human reconstruction and the GT surface, widely used in previous works~\cite{saito2019pifu,saito2020pifuhd,jiang2020bcnet}.
Before measurement, we rigidly align global rotation, scale, and translation of the reconstruction to the GT surface.
The alignment is not trivial as there are no semantically matching pairs between the reconstruction and GT surface.
For the alignment, we assume that the same human parts (\textit{e.g.}, shoulders and knees) of both the reconstruction and the GT surface are projected to the same pixel in the image.
We rasterize both the reconstruction and GT surface and pair two vertices (\textit{i.e.}, one from the reconstruction and the other from the GT surface) that correspond to the same pixel.
Based on the matching pairs, we align the reconstruction and measure the Chamfer distance (CD) in millimeter.
More details of CD metric is described in the supplementary material.

\setlength{\tabcolsep}{4pt}
\begin{table}[t]
\centering
\caption{
    CD comparison with state-of-the-art methods on 3DPW.
    Methods with $\ast$ and $\dagger$ additionally use human segmentation and cloth segmentations as an input for the inference, respectively.
}
\label{table:comparison_cd}
\def\arraystretch{1.3}
\begin{tabular}{c|c}
\hline
Methods                                      & CD $\downarrow$\\ \hline
PIFuHD~\cite{saito2020pifuhd}  & 137.50        \\
BCNet~\cite{jiang2020bcnet} & 118.75         \\
ICON~\cite{xiu2022icon} & 75.00 \\
PIFu~\cite{saito2019pifu}$\ast$ & 67.25         \\
SMPLicit fits~\cite{corona2021smplicit}$\dagger$  & 45.66 \\ 
\textbf{ClothWild (Ours)} & \textbf{40.34}\\
\hline
\end{tabular}
\end{table}
\setlength{\tabcolsep}{1.4pt}

\noindent\textbf{Body-cloth correspondence (BCC).}
Body-cloth correspondence (BCC) is a proportion of 3D points that have correct cloth types on the T-posed naked human body surface.
It only considers 3D cloth predictions, excluding 3D human pose predictions.
With cloth-to-body mapping described in Section~\ref{sec:dp_loss}, we map 2D points in a GT cloth segmentation for one cloth label to 3D points on the T-posed human body surface with the cloth-to-body mapping.
We consider 3D points are correctly covered ones if the distance between them and reconstructed cloth is shorter than 3 cm.
We calculate the proportion of the correctly covered points and average the proportions for all cloth labels (\textit{e.g.}, upper cloth, pants, or non-cloth).
This metric is used for methods that support cloth-to-body mapping based on the T-posed naked body of SMPL.
For example, as PIFu~\cite{saito2019pifu}, PIFuHD~\cite{saito2020pifuhd}, and ICON~\cite{xiu2022icon} do not tell us cloth type which points belong, we could not evaluate them using the BCC metric.

\subsection{Comparison with state-of-the-art methods}
~\label{sec:compare_sota}
We compare our ClothWild with recent 3D clothed human reconstruction methods: PIFu~\cite{saito2019pifu}, PIFuHD~\cite{saito2020pifuhd}, ICON~\cite{xiu2022icon}, BCNet~\cite{jiang2020bcnet}, and the fitting framework of SMPLicit~\cite{corona2021smplicit}.
For the inference, PIFu requires human segmentation, and the fitting framework of SMPLicit requires SMPL parameter and cloth segmentations.
We provide human segmentation, obtained by Mask R-CNN~\cite{he2017mask}, to PIFu.
The same SMPL parameters of our BodyNet and cloth segmentations from SCHP~\cite{li2020self} are provided to the SMPLicit fitting framework.
All of their results are obtained by using their officially released codes and pre-trained weights.
Please note that the fitting framework of SMPLicit fits their cloth latent codes to the cloth segmentations.
We call the fitting results of their framework as SMPLicit fits.

\setlength{\tabcolsep}{4pt}
\begin{table}[t]
\centering
\caption{
    BCC comparison with state-of-the-art methods on MSCOCO.
    Methods with $\dagger$ additionally use cloth segmentations as an input for the inference.
}
\label{table:comparison_bcc}
\def\arraystretch{1.3}
\scalebox{1.0}{
\begin{tabular}{c|c|c|c|c}
\hline
Methods  &  upper body & lower body & non-cloth & average  \\ \hline
BCNet~\cite{jiang2020bcnet}                         & 0.415     & 0.729     & 0.800     & 0.648          \\
SMPLicit fits~\cite{corona2021smplicit}$\dagger$    & 0.645     & 0.493     & \textbf{0.961}     & 0.700          \\ 
\textbf{ClothWild (Ours)}                      & \textbf{0.830}     & \textbf{0.820}     & 0.887     & \textbf{0.846} \\
\hline
\end{tabular}
}
\end{table}
\setlength{\tabcolsep}{1.4pt}

\noindent\textbf{Qualitative results.}
Fig.~\ref{fig:comparison_sota} shows that our ClothWild produces much better reconstruction results than previous state-of-the-art 3D clothed human reconstruction methods on MSCOCO~\cite{lin2014microsoft}.
PIFu~\cite{saito2019pifu}, PIFuHD~\cite{saito2020pifuhd}, and ICON~\cite{xiu2022icon} suffer from undesirable results on in-the-wild images with diverse human poses and appearances, especially in the occluded human scene.
ICON produces better results than PIFu and PIFuHD; however, it still suffers from missing human parts.
On the other hand, our ClothWild successfully reconstructs invisible parts of clothed humans even when the input human is occluded or truncated.
Like PIFu, PIFuHD, and ICON, BCNet~\cite{jiang2020bcnet} also requires 3D scans for the training.
Accordingly, it is trained on synthetic datasets without in-the-wild images, the reason for weak results on in-the-wild images.
SMPLicit fits~\cite{corona2021smplicit} does not utilize image features, which provide contextual information of occluded and truncated human parts.
Hence, it cannot reconstruct clothes at occluded or truncated human parts.
In addition, the depth ambiguity and pixel-level misalignment of their silhouette-based loss function make the fitting attain the incorrect cloth styles.
Our ClothWild resolves such issues by utilizing image features and the DensePose-based loss function.

\noindent\textbf{Quantitative results.}
Table~\ref{table:comparison_cd} shows that ClothWild achieves the best CD on 3DPW~\cite{lin2014microsoft}.
Also, Table~\ref{table:comparison_bcc} shows that ClothWild produces much better BCC than previous methods on MSCOCO~\cite{lin2014microsoft}.
Unlike ours, BCNet~\cite{jiang2020bcnet} handles only three cloth types (\textit{i.e.}, upper cloth, pants, and skirt).
Hence, we categorize upper cloth and coat into the upper body and pants and skirts into the lower body for a fair comparison.
Then, we calculate BCC only for such clothes and exclude shoes.
As the BCC evaluates only predicted 3D clothes without considering 3D human poses, the comparisons show that our ClothWild predicts the cloth styles of the image accurately, such as sleeveless and length of upper cloth and pants.

\setlength{\tabcolsep}{4pt}
\begin{table}[t]
\centering
\caption{
    Running time (seconds per image) comparisons.
}
\label{table:running_time}
\def\arraystretch{1.4}
\begin{tabular}{cccc}
\hline
\begin{tabular}{c} PIFuHD~\cite{saito2020pifuhd}\\ \fontsize{6}{6}\selectfont{(CVPR 20)} \end{tabular}  & \begin{tabular}{c} SMPLicit fits~\cite{corona2021smplicit} \\ \fontsize{6}{6}\selectfont{(CVPR 21)} \end{tabular} & \begin{tabular}{c} ICON~\cite{xiu2022icon} \\  \fontsize{6}{6}\selectfont{(CVPR 22)} \end{tabular} & \begin{tabular}{c} \textbf{ClothWild} \\  \textbf{(Ours)} \end{tabular} \\ \hline
20.43 & 105.43 & 87.88 & \textbf{10.21}\\
\hline
\end{tabular}
\end{table}
\setlength{\tabcolsep}{1.4pt}

\noindent\textbf{Running time.}
Table~\ref{table:running_time} shows that ours takes the shortest computational time to process a single image than recently presented works.
The running times are measured in the same environment with Intel Xeon Gold 6248R CPU and NVIDIA RTX 2080 Ti GPU.
We exclude pre-processing stages, such as 3D body pose estimation, human segmentation, and cloth segmentations.
The fitting framework of SMPLicit takes a much longer time, although it is based on the same cloth generative model~\cite{corona2021smplicit} as ours.
This is because the fitting framework of SMPLicit forwards the cloth generative model about 200 iterations to fit its cloth latent codes to cloth segmentations.
ICON suffers from a similar problem as it iteratively fits initial results based on normal maps and silhouettes.
In contrast, our ClothWild performs the feed-forward only a single time.
The above methods, including ours, require Marching Cubes to obtain final 3D geometry, one of the main bottlenecks of the running time.
The running time of ClothWild's each component will be reported in the supplementary material.

\subsection{Ablation study}
\begin{figure}[t]
\centering
\includegraphics[width=1.0\linewidth]{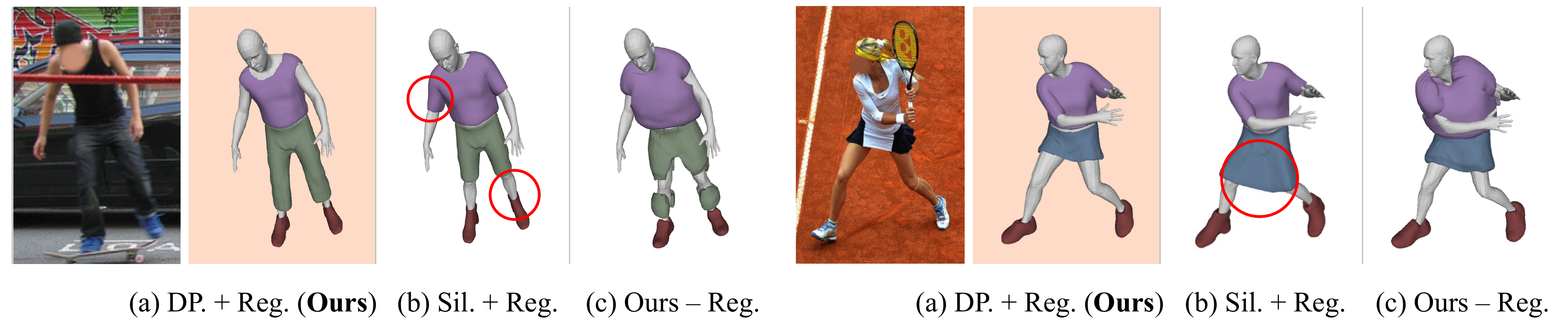}
\caption{
Comparison of three different loss configurations:
(a) our proposed loss configuration,
(b) replacing our DensePose-based loss with the silhouette loss, and
(c) removing the regularization loss.
}
\label{fig:ablation_study}
\end{figure}

\noindent\textbf{Effectiveness of DensePose-based loss.}
~\label{sec:ablation_densepose_loss}
The top block of Table~\ref{table:ablation_study} shows that the proposed DensePose-based loss has a vital role in properly learning 3D clothed humans.
Fig.~\ref{fig:ablation_study} (a) and (b) additionally shows the effectiveness of our DensePose-based loss compared to previous silhouette loss~\cite{corona2021smplicit}.
The silhouette loss enforces 3D cloth reconstruction results to be projected onto the 2D cloth segmentations.
The top block of the table shows that using the DensePose-based loss achieves better BCC than using the silhouette loss.
The silhouette loss suffers from depth ambiguity and pixel-misalignment, as described in Section~\ref{sec:introduction}.
On the other hand, the proposed DensePose-based suffers much less from the depth ambiguity and is free from the misalignment issue as it designates accurate 3D points around the human body surface.
Hence, it has significant benefits to learning precise 3D clothes.

\setlength{\tabcolsep}{4pt}
\begin{table}[t]
\centering
\caption{
    BCC comparison among different loss configurations.
}
\label{table:ablation_study}
\def\arraystretch{1.3}
\begin{tabular}{ccc|c}
\hline
DensePose & Silhouette & Regularization& BCC $\uparrow$\\ \hline
\multicolumn{3}{l|}{\textbf{* Effectiveness of DensePose-based loss}} & \\
\xmark & \cmark & \cmark & 0.644\\
\cmark & \cmark & \cmark & 0.684 \\
\cmark & \xmark & \cmark & \textbf{0.689 (Ours)} \\ \hline
\multicolumn{3}{l|}{\textbf{* Effectiveness of regularization loss}} & \\
\cmark & \xmark & \xmark & 0.381 \\
\cmark & \xmark & \cmark & \textbf{0.689 (Ours)} \\
\hline
\end{tabular}
\end{table}
\setlength{\tabcolsep}{1.4pt}

\noindent\textbf{Effectiveness of regularization loss.}
~\label{sec:ablation_regularization_loss}
The bottom block of Table~\ref{table:ablation_study} shows the effectiveness of the regularization loss.
Fig.~\ref{fig:ablation_study} (b) and (c) additionally demonstrates the effectiveness.
The regularization loss makes predicted clothes be in the latent space of the 3D cloth model.
It is necessary because 2D supervision targets, such as cloth segmentations, only provide information of partial areas of clothes.
Such limited information can lead our ClothWild to produce implausible clothes, such as overly thick cloth and torn cloth.
The regularization loss prevents such improper learning and encourages reconstructing reasonable 3D clothes, despite only information of partial areas of clothes.   
\section{Conclusion}
We propose ClothWild, a 3D clothed human reconstruction framework that produces significantly robust results from in-the-wild images.
For the robustness to the domain gap between synthesized and in-the-wild datasets, we propose a weakly supervised pipeline and a DensePose-based loss function.
As a result, our ClothWild outperforms previous 3D clothed human reconstruction methods on in-the-wild images.

\noindent\textbf{Acknowledgements.}
This work was supported in part by IITP grant funded by the Korea government (MSIT) [No. 2021-0-01343, Artificial Intelligence Graduate School Program (Seoul National University), No.2022-0-00156], and in part by the Bio \& Medical Technology Development Program of NRF funded by the Korean government (MSIT) [No. 2021M3A9E4080782].

\clearpage

\begin{center}
\textbf{\large Supplementary Material \textit{for} \\ \vspace{2mm}
\large{``3D Clothed Human Reconstruction in the Wild"}}
\end{center}

In this supplementary material, we present more technical details and additional experimental results that could not be included in the main manuscript due to the lack of space.

\section{Controlling reconstruction results}
\begin{figure} [h!]
\centering
\includegraphics[width=1.0\linewidth]{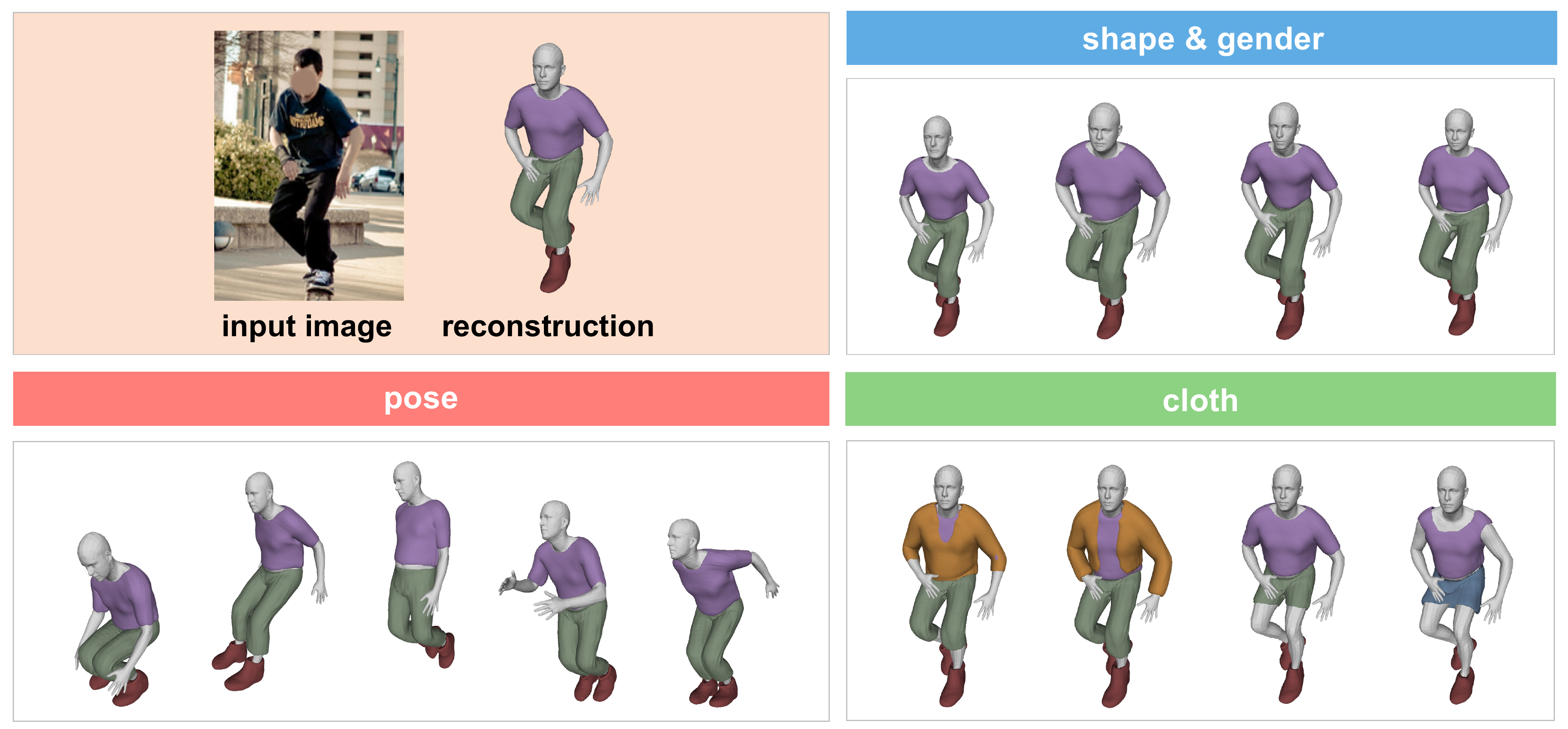}
   \caption{
   An example for controlling the reconstructed 3D clothed human.
   Our reconstruction results are editable for pose, shape, gender, and cloth style.
   }
\label{fig:controllable}
\end{figure}

Our ClothWild has additional strength that the reconstructed results can be modified easily for other pose, shape, gender, and cloth style.
In our framework, the pose, shape, gender, and clothes are disentangled by predicting their latent codes separately.
Thus, we can edit the predicted latent codes as we want and forward them into the 3D cloth and human models (\textit{i.e.}, SMPLicit~\cite{corona2021smplicit} and SMPL~\cite{loper2015smpl}) of our framework.
Fig.~\ref{fig:controllable} shows an example of controlling a reconstructed 3D clothed human in three different categories.
First, we can animate the reconstruction result by editing pose parameters represented by 3D rotations of human body joints (red part of the figure).
Second, by modifying the shape and gender, we can change the naked human body of the reconstructed 3D clothed human (blue part of the figure).
Third, clothes are represented by cloth existence scores and cloth latent codes of our framework, and we can change clothes by applying different cloth existence scores and cloth latent codes of other clothed humans (green part of the figure).

\section{Running time of ClothWild's components}
Table~\ref{table:running_time} shows that the ClothNet has an extremely short running time compared to other components in our framework.
Furthermore, a significant portion of the running time is occupied by components that generate 3D clothed humans from the latent codes, not our ClothNet.
In our experiment in Table~\ref{table:running_time}, the ClothNet and the SMPLicit are running on GPU (\textit{i.e.}, RTX 2080 Ti), and the Marching Cubes and the Pose deformation are running on CPU (\textit{i.e.}, Intel Xeon Gold 6248R) following the convention.

\setlength{\tabcolsep}{4pt}
\begin{table}
\centering
\caption{
    Running time of each component of our ClothWild, where the unit of time is second.
}
\label{table:running_time}
\def\arraystretch{1.4}
\scalebox{1.0}{
\begin{tabular}{cccc|c}
\hline
ClothNet & SMPLicit & Marching Cubes & Pose deformation & \textbf{Total} \\ \hline
0.01 & 1.08 & 3.82 & 5.10 & \textbf{10.21} \\
\hline
\end{tabular}
}
\end{table}
\setlength{\tabcolsep}{1.4pt}

\section{Detail of sampling strategy for query point selection}
\begin{figure}[t]
\includegraphics[width=1.0\linewidth]{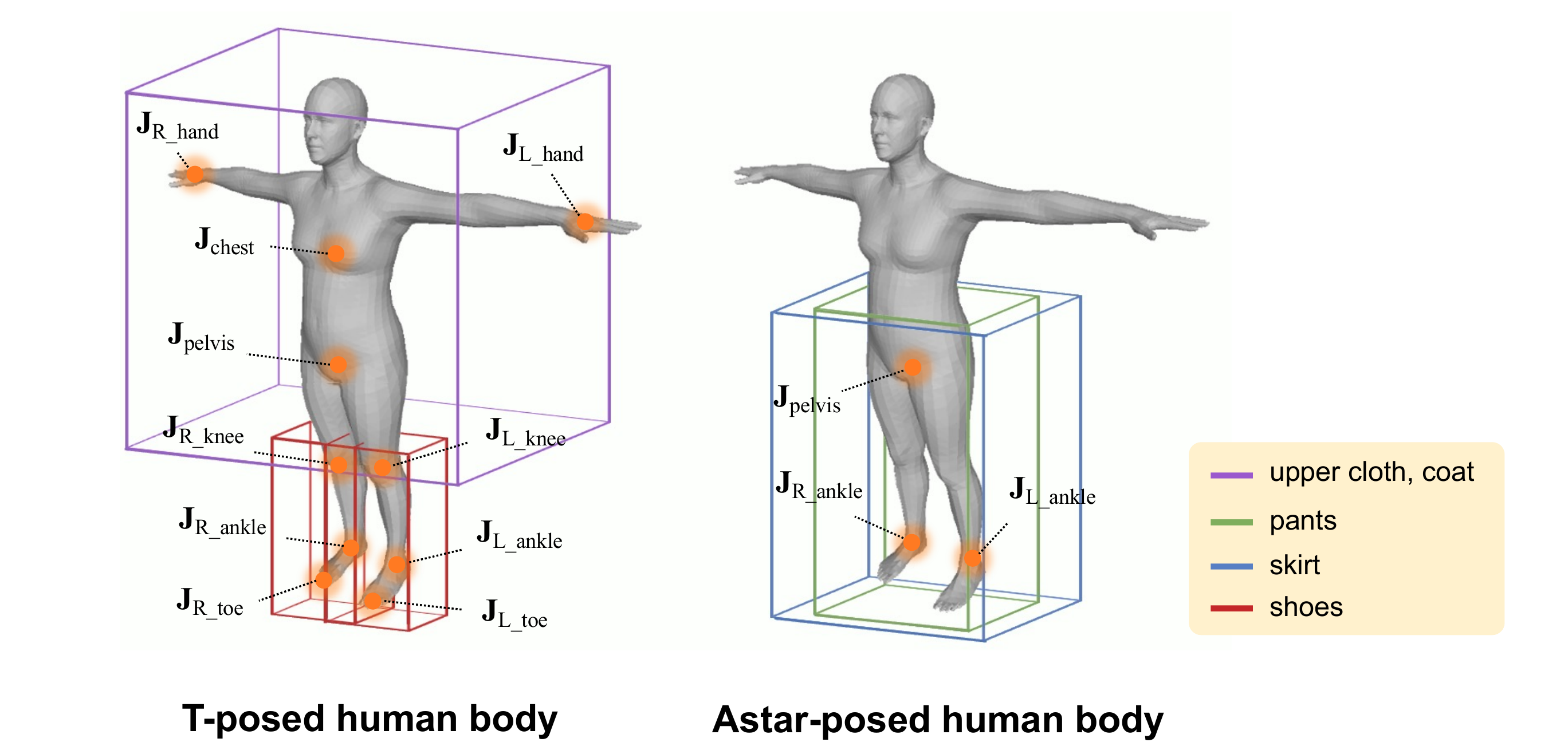}
\centering
\caption{
Illustration of 3D bounding boxes for query point selection.
}
\label{fig:query_selection}
\end{figure}

In the query point selection step, we uniformly sample 3D query points at a resolution of $21\times21\times21$ from a 3D bounding box for each cloth.
The 3D bounding boxes are determined by 3D joints coordinates of the T-posed human body, where the 3D joints are defined in SMPL~\cite{loper2015smpl}.
The 3D bounding box for each cloth and the joints is illustrated in Fig.~\ref{fig:query_selection}.
To the formal description, we define several notations.
We denote 3D coordinates of a joint as $\mathbf{J}_{\bullet}$, and its $x$, $y$, and $z$ value of the 3D coordinates are denoted as $J_{\bullet}^x$, $J_{\bullet}^y$, and $J_{\bullet}^z$.
Additionally, we denote minimum and maximum of $z$ values of T-posed human body vertices as $v_{\text{min}}^z$ and $v_{\text{max}}^z$.
The 3D bounding box for each cloth is represented as a corner representation $[x_{\text{min}}, y_{\text{min}}, z_{\text{min}}, x_{\text{max}}, y_{\text{max}}, z_{\text{max}}]$, where $\bullet_{\text{min}}$ and $\bullet_{\text{max}}$ are the minimum and maximum values of each coordinate at eight corners of the 3D bounding box.
The 3D bounding boxes follow:   
\begin{equation}
\begin{split}
    \scriptsize [x_{\text{min}}, y_{\text{min}}, z_{\text{min}}, x_{\text{max}}, y_{\text{max}}, z_{\text{max}}] = \qquad\qquad\qquad\qquad\qquad\quad\\
     \scriptsize [J_{\text{R\_hand}}^x, 2J_{\text{pelvis}}^y-J_{\text{chest}}^y, 1.25v_{\text{min}}^z-0.25v_{\text{max}}^z, \qquad\qquad\;\\
     \scriptsize J_{\text{L\_hand}}^x, 3J_{\text{chest}}^y-2J_{\text{pelvis}}^y, 1.25v_{\text{max}}^z-0.25v_{\text{min}}^z], 
\end{split}
\end{equation}
for upper cloth and coat,
\begin{equation}
\begin{split}
    \footnotesize [x_{\text{min}}, y_{\text{min}}, z_{\text{min}}, x_{\text{max}}, y_{\text{max}}, z_{\text{max}}] = \qquad\qquad\qquad\qquad\qquad\;\;\\
     \footnotesize [J_{\text{L\_ankle}}^x - 0.15, 1.75J_{\text{L\_ankle}}^y - 0.75J_{\text{L\_knee}}^y, 0.5J_{\text{L\_ankle}}^z + 0.5J_{\text{L\_toe}}^z - 0.25, \\
     \footnotesize J_{\text{L\_ankle}}^x + 0.15, 0.25J_{\text{L\_ankle}}^y + 0.75J_{\text{L\_knee}}^y, 0.5J_{\text{L\_ankle}}^z + 0.5J_{\text{L\_toe}}^z + 0.25], 
\end{split}
\end{equation}
for left shoe, and 
\begin{equation}
\begin{split}
    \scriptsize [x_{\text{min}}, y_{\text{min}}, z_{\text{min}}, x_{\text{max}}, y_{\text{max}}, z_{\text{max}}] = \qquad\qquad\qquad\qquad\qquad\;\;\\
     \scriptsize [J_{\text{R\_ankle}}^x - 0.15, 1.75J_{\text{R\_ankle}}^y - 0.75J_{\text{R\_knee}}^y, 0.5J_{\text{R\_ankle}}^z + 0.5J_{\text{R\_toe}}^z - 0.25, \\
     \scriptsize J_{\text{R\_ankle}}^x + 0.15, 0.25J_{\text{R\_ankle}}^y + 0.75J_{\text{R\_knee}}^y, 0.5J_{\text{R\_ankle}}^z + 0.5J_{\text{R\_toe}}^z + 0.25], 
\end{split}
\end{equation}
for right shoe.
The 3D bounding boxes for pants and a skirt are formed based on the Astar-posed human body, which legs are slightly wider than the T-posed human body, as follows:
\begin{equation}
\begin{split}
    \scriptsize [x_{\text{min}}, y_{\text{min}}, z_{\text{min}}, x_{\text{max}}, y_{\text{max}}, z_{\text{max}}] = \qquad\qquad\qquad\qquad\qquad\\
     \scriptsize [2.3J_{\text{R\_ankle}}^x - 1.3J_{\text{pelvis}}^x, 1.1J_{\text{R\_ankle}}^y - 0.1J_{\text{pelvis}}^y, 1.25v_{\text{min}}^z - 0.25v_{\text{max}}^z, \\
     \scriptsize 3.3J_{\text{pelvis}}^x - 2.3J_{\text{R\_ankle}}^x, 1.1J_{\text{pelvis}}^y - 0.1J_{\text{R\_ankle}}^y, 1.25v_{\text{max}}^z - 0.25v_{\text{min}}^z], 
\end{split}
\end{equation}
for pants, and
\begin{equation}
\begin{split}
    \scriptsize [x_{\text{min}}, y_{\text{min}}, z_{\text{min}}, x_{\text{max}}, y_{\text{max}}, z_{\text{max}}] = \qquad\qquad\qquad\qquad\qquad\\
     \scriptsize [3J_{\text{R\_ankle}}^x - 2J_{\text{pelvis}}^x, 1.1J_{\text{R\_ankle}}^y - 0.1J_{\text{pelvis}}^y, 1.25v_{\text{min}}^z - 0.25v_{\text{max}}^z, \quad \\
     \scriptsize 4J_{\text{pelvis}}^x - 3J_{\text{R\_ankle}}^x, 1.1J_{\text{pelvis}}^y - 0.1J_{\text{R\_ankle}}^y, 1.25v_{\text{max}}^z - 0.25v_{\text{min}}^z], 
\end{split}
\end{equation}
for skirt.

\begin{figure}
\includegraphics[width=1.0\linewidth]{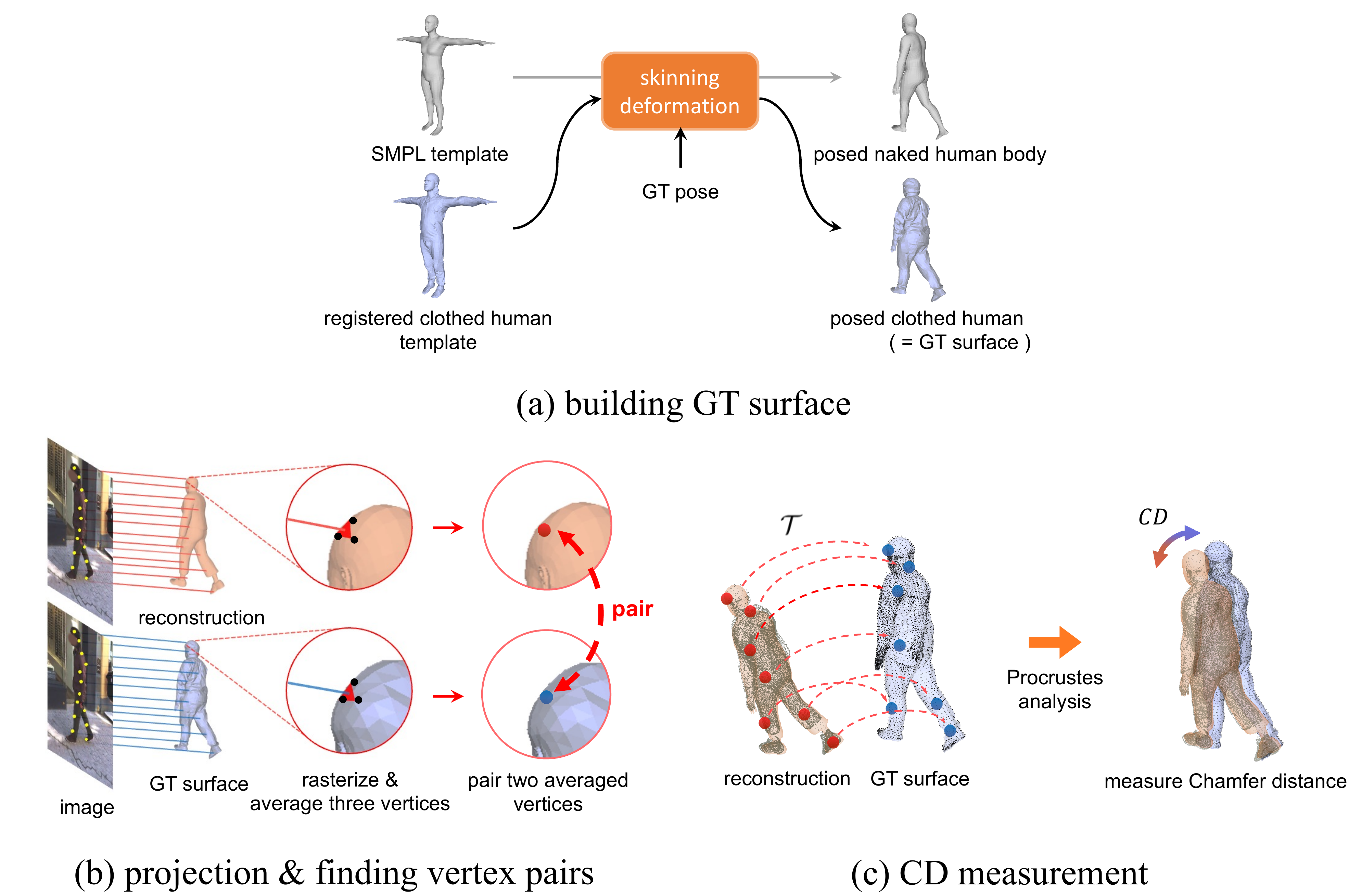}
\centering
   \caption{
   Detailed illustration of the three steps of the Chamfer distance evaluation.
   }
\label{fig:chamfer_description}
\end{figure}

\section{Detail of Chamfer distance metric}
This section details the Chamfer distance (CD) evaluation metric in three steps: building GT surface, projection \& finding vertex pairs, and CD measurement.

\noindent\textbf{Building GT surface.}
In 3DPW~\cite{von2018recovering}, used for evaluation in our work, T-posed clothed human meshes are registered to each subject's 3D scan.
The registered 3D clothed human meshes have the same mesh topology as that of SMPL body mesh.
We deform the registered 3D clothed humans with GT SMPL pose parameters following the same skinning deformation of SMPL.
We use the posed clothed human as GT surface to measure CD with reconstruction result.

\noindent\textbf{Projection \& finding vertex pairs.}
Before measuring CD, we rigidly align global rotation, scale, and translation of the reconstruction to the GT surface.
For the alignment, we find semantically matching vertex pairs between the reconstruction and the GT surface based on each 2D projection.
First, we project both the reconstruction and GT surface into the input image and rasterize them.
By the rasterization, we obtain face index maps of their meshes, where each pixel of a face index map represents a visible face index among mesh faces projected to that pixel location.
Second, we average three vertices that make up each face for both the reconstruction and GT surface.
Finally, we pair the averaged two vertices (\textit{i.e.}, one from the reconstruction and the other from the GT surface) that correspond to the same pixel location. 
With these processes, we obtain semantic matching vertex pairs that project to the same pixel of the input image.

\noindent\textbf{CD measurement.}
Based on the semantic matching vertex pairs, we construct a rigid transformation matrix $\mathcal{T}$ that transforms the first vertices to the second vertices of the pairs.
Using the rigid transformation matrix $\mathcal{T}$, we align all of the reconstruction vertices to the GT surface.
After alignment, we measure CD between the aligned reconstruction and the GT surface.

\section{Supervision on cloth existence}
\label{sec:cloth_existence_supervision}
GT existence of each cloth is set to \texttt{True} if a GT segmentation of the cloth exists in the input image.
As cloth segmentations are defined only inside of the image, naively setting the existence to \texttt{False} when the segmentation does not exist in the input image can be wrong.
For example, if a human is wearing pants and the pants are not included in the input image due to the truncation, the naive solution sets the existence of the pants to \texttt{False}, which is wrong.

Instead, we use DensePose to distinguish 1) a human is not wearing a cloth and 2) a human is possibly wearing a cloth but not included in the input image.
In the DensePose, human part patches are defined, such as head, torso, arm, and so on. 
We correspond the upper clothes and the coat into torso and arm parts, the pants and the skirt into leg parts, and the shoes into foot parts.
The existence of each cloth is set to \texttt{False} if 1) the cloth's human part patches are included in the input image and 2) the cloth's segmentation does not exist in the input image.
If the cloth's human part patches are not included in the input image, we do not supervise the predicted cloth existence score for the cloth.

\section{Evaluation on cloth existence and gender}
We evaluate our ClothWild in cloth existence and gender on the MSCOCO~\cite{lin2014microsoft} validation set.
GT cloth existences are obtained cloth segmentation annotations~\cite{gong2017look} and DensePose in the same way as described in Section~\ref{sec:cloth_existence_supervision}.
For gender, since there is no dataset containing GT gender annotations for MSCOCO, we acquire gender annotations by running Homogenus~\cite{pavlakos2019expressive} and use its predictions as pseudo-GTs.
Measuring accuracy with the obtained GTs, the accuracy of cloth existence and gender are 0.848 and 0.918, respectively.

\section{Limitations}
\begin{figure}[h!]
\includegraphics[width=1.0\linewidth]{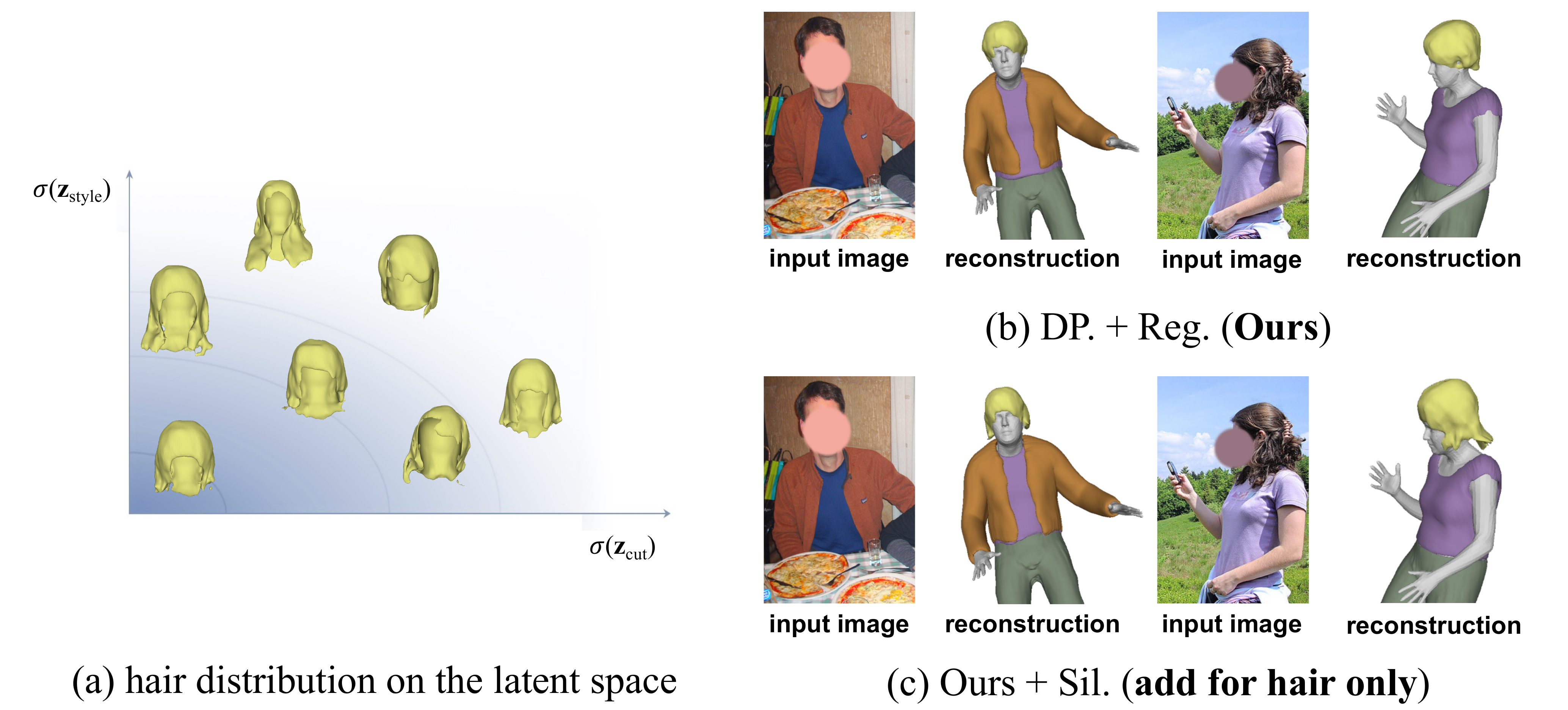}
\centering
   \caption{
   (a) 3D hair distribution according to the standard deviations of the latent codes.
   (b) Hair reconstruction examples with our loss configuration.
   (c) Hair reconstruction examples when adding the silhouette loss for hair only.
   }
\label{fig:hair_reconstruction}
\end{figure}

\noindent\textbf{Hair reconstruction.}
Although the cloth generative model, SMPLicit ~\cite{corona2021smplicit}, also supports 3D hair generation, there is a limitation to reconstructing hair in our framework for the following two reasons.
First, the SMPLicit cannot cover a wide variety of hair because it is trained mostly from a small set of women's hair.
Fig.~\ref{fig:hair_reconstruction}(a) shows that SMPLicit's hair outputs are not diverse and mostly long hair, biased to women's long hair.
The hairs of the figure are produced from randomly sampled two Gaussian codes (\textit{i.e.}, $\textbf{z}_{\text{style}}$ and $\textbf{z}_{\text{cut}}$) with various standard deviations following the model design of the SMPLicit.
Second, a large part of the hair is quite distant from the human body surface.
As our DensePose-based loss function supervises 3D query points close to the human body surface, a large part of the hair is not covered by the DensePose-based loss.
Although we use silhouette loss to supervise 3D query points far from the surface, learning 3D hair is not done properly due to the depth ambiguity of 2D supervision targets.
Figs.~\ref{fig:hair_reconstruction}(b) and~\ref{fig:hair_reconstruction}(c) show that the reconstructed hair with our framework is unnatural, especially for long hair.
Due to these difficulties, learning 3D hair is one of the challenges to be solved.

\begin{figure} [t!]
\includegraphics[width=1.0\linewidth]{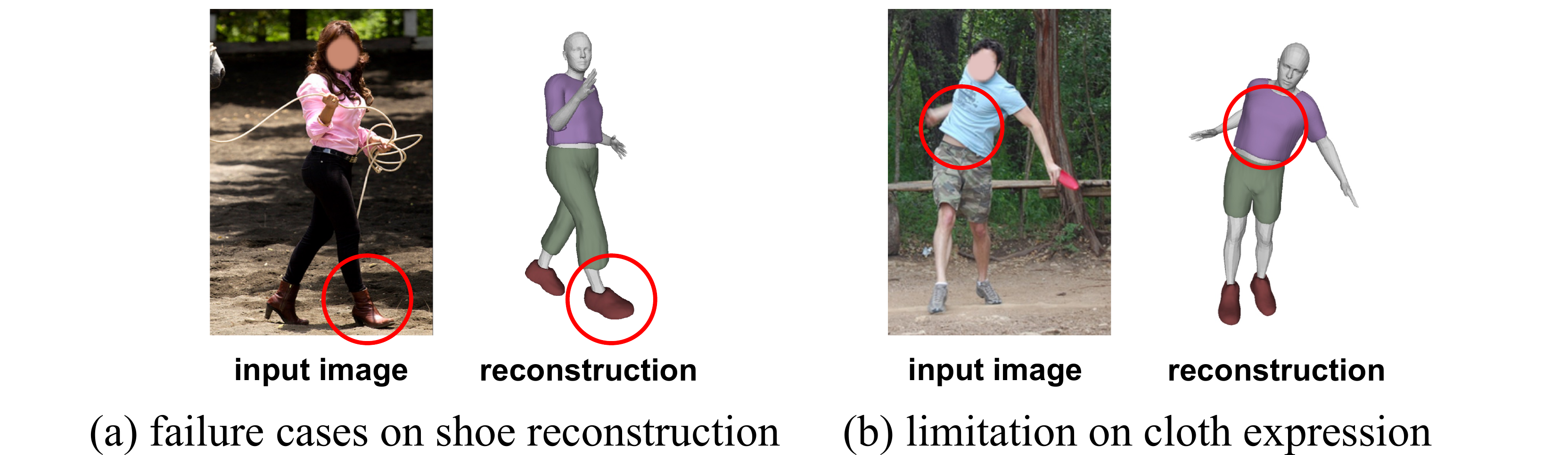}
\centering
   \caption{
   Limitations of our framework: shoe reconstruction and expression power.
   }
\label{fig:limitation}
\end{figure}

\noindent\textbf{Shoes reconstruction.}
Fig.~\ref{fig:limitation}(a) shows that our framework often inadequately reconstructs several shoes, such as high heels.
We guess the reason is that such shoes have complex geometry, while they do not have many pixels in the cloth segmentations.
Therefore, the shoe segmentations are not very informative for learning shoes with complex geometry.

\noindent\textbf{Expression power.}
There is a limitation to represent cloth details (\textit{e.g.}, wrinkles), as shown in Fig.~\ref{fig:limitation}(b).
The reason is that the expression power of our framework depends on the cloth generative model.
Most of current cloth generative models~\cite{bhatnagar2019multi,jiang2020bcnet,ma2020learning,bertiche2020cloth3d,patel2020tailornet}, including SMPLicit~\cite{corona2021smplicit} of our framework, have difficulty in embedding delicate cloth geometry (\textit{e.g.}, wrinkles) and in the cloth latent space.
Therefore, we think improving the expression power of cloth generative models should be a future research direction.

\clearpage

\begin{figure} [b!]
\includegraphics[width=1.0\linewidth]{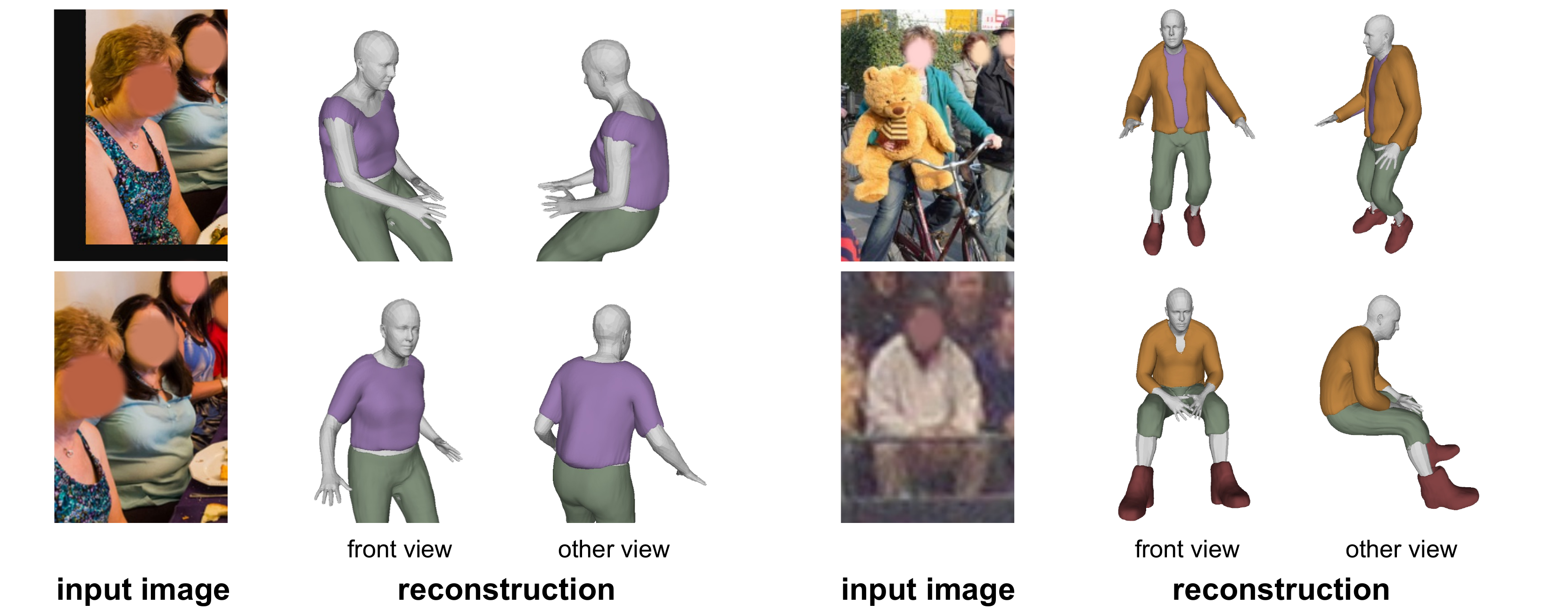}
\centering
   \caption{
   Qualitative results on challenging cases of MSCOCO~\cite{lin2014microsoft} validation set.
   Colors of reconstructed 3D clothes are manually assigned to represent cloth types.
   }
\label{fig:challenging_cases}
\end{figure}

\section{More qualitative results}
\begin{figure*}
\begin{center}
\includegraphics[width=1.0\linewidth]{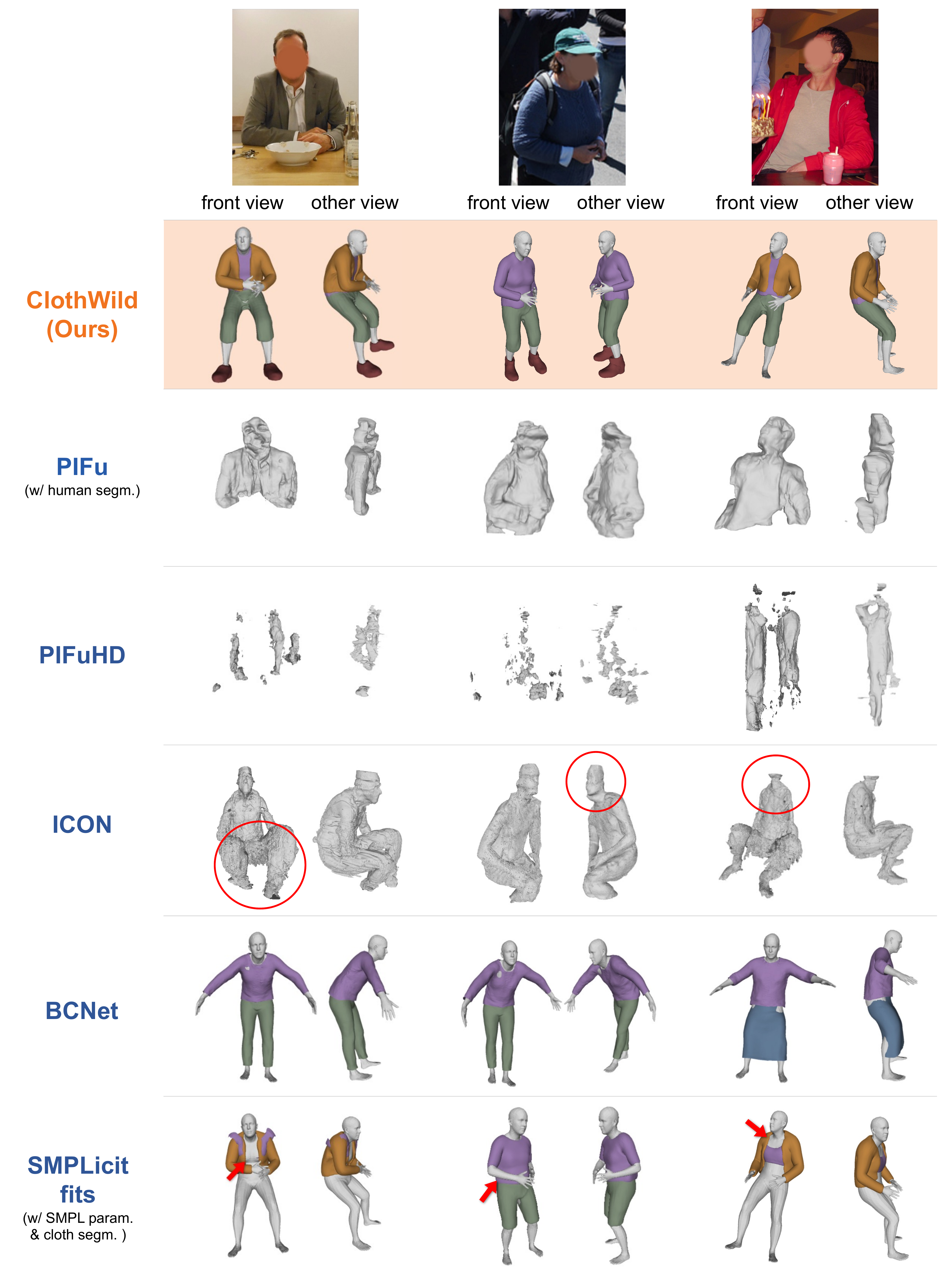}
\end{center}
   \caption{
    Qualitative result comparisons on MSCOCO~\cite{lin2014microsoft} validation set.    
    PIFu additionally uses human segmentation obtained from Mask R-CNN~\cite{he2017mask} for reconstruction.
    SMPLicit fits use SMPL parameter and cloth segmentations obtained from Pose2Pose~\cite{Moon_2022_CVPRW_Hand4Whole} and SCHP~\cite{li2020self}, respectively. 
    Colors of reconstructed 3D clothes are manually assigned to represent cloth types.
}
\label{fig:more_qualitative_1}
\end{figure*}

\begin{figure*}
\begin{center}
\includegraphics[width=1.0\linewidth]{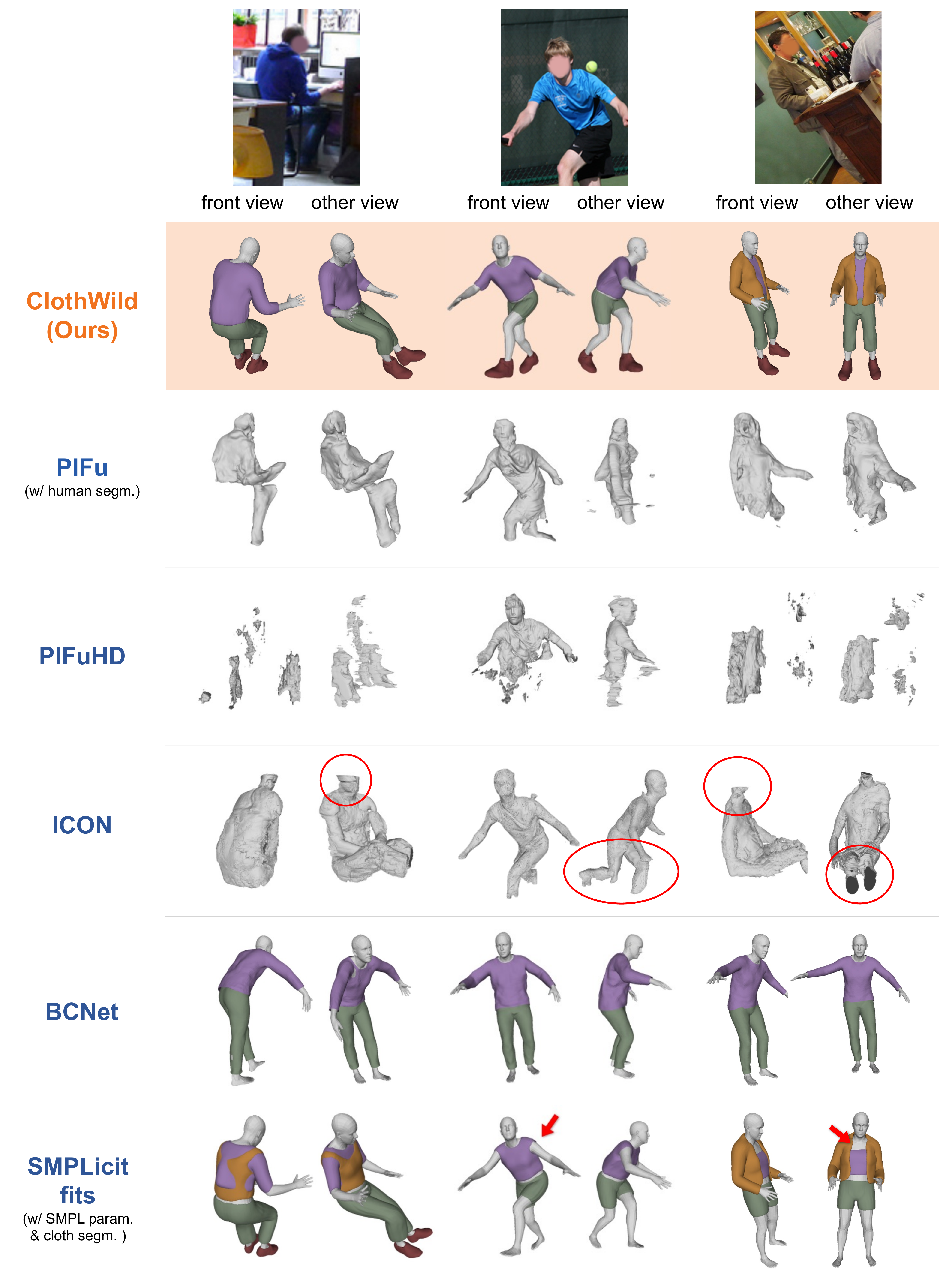}
\end{center}
   \caption{
    Qualitative result comparisons on MSCOCO~\cite{lin2014microsoft} validation set.
    PIFu additionally uses human segmentation obtained from Mask R-CNN~\cite{he2017mask} for reconstruction.
    SMPLicit fits use SMPL parameter and cloth segmentations obtained from Pose2Pose~\cite{Moon_2022_CVPRW_Hand4Whole} and SCHP~\cite{li2020self}, respectively. 
    Colors of reconstructed 3D clothes are manually assigned to represent cloth types.
}
\label{fig:more_qualitative_2}
\end{figure*}

We provide more qualitative result comparisons on the MSCOCO~\cite{lin2014microsoft} validation set.
Fig.~\ref{fig:challenging_cases} shows that our ClothWild performs well on extremely challenging cases, such as overlaps between people (upper left and lower left), occlusions (upper right), and from a low-resolution image (lower right).
Figs.~\ref{fig:more_qualitative_1} and ~\ref{fig:more_qualitative_2} show our ClothWild produces far better results from in-the-wild images compared to previous state-of-the-art 3D clothed human reconstruction methods.

\clearpage

\bibliographystyle{splncs04}
\bibliography{bib}

\end{document}